\pdfoutput=1

\documentclass[11pt]{article}

\usepackage{acl}

\usepackage{times}
\usepackage{latexsym}
\usepackage{mathtools}
\usepackage{amsmath,amssymb,mathtools}

\usepackage[ruled,vlined]{algorithm2e}

\usepackage{hyperref}
\usepackage{subcaption}
\usepackage{fontawesome5}

\usepackage[T1]{fontenc}

\usepackage[utf8]{inputenc}

\usepackage{amsmath} 

\usepackage{microtype}

\usepackage{inconsolata}

\usepackage{graphicx}

\title{\raisebox{-0.5em}{\includegraphics[height=1.8em]{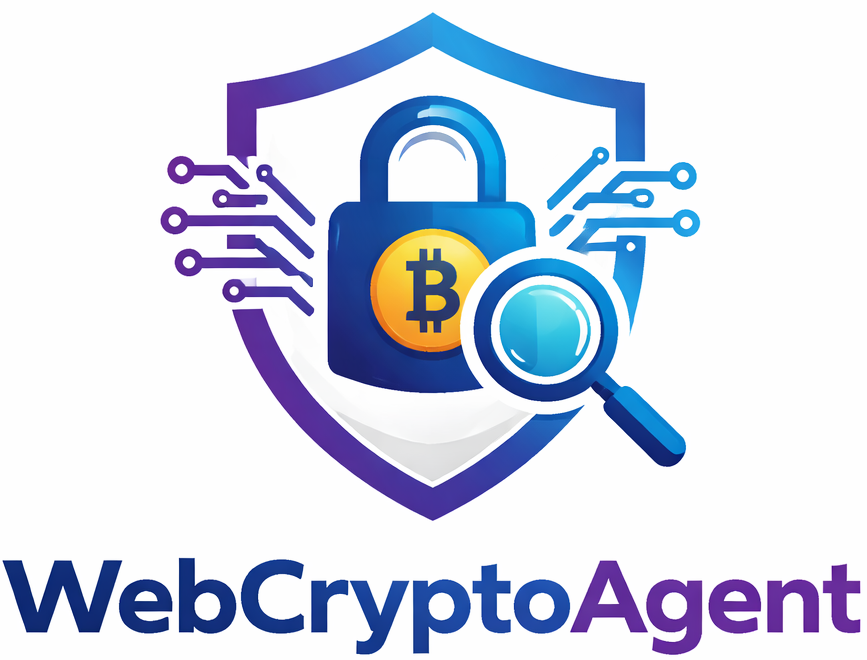}}~WebCryptoAgent: Agentic Crypto Trading with Web Informatics}

\author{Ali Kurban$^{1*}$\quad
Wei Luo$^{2*}$\quad
Liangyu Zuo$^{1*}$\quad
Zeyu Zhang$^{3\dag}$\\
\textbf{Renda Han}$^{4}$\quad
\textbf{Zhaolu Kang}$^{3}$\quad
\textbf{Hao Tang}$^{3}$\quad
\textbf{Yang Zhao}$^{5\ddag}$\quad\\
\vspace{0.2cm}
$^1$AI Geeks\quad
$^2$CUHK\quad
$^3$Peking University\quad
$^4$TJU\quad
$^5$La Trobe\\
\small $^*$Equal contribution. $^\dag$Project lead. $^\ddag$Corresponding author: y.zhao2@latrobe.edu.au.}

\begin{document}
\maketitle
\begin{abstract}
Cryptocurrency trading increasingly depends on timely integration of heterogeneous web information and market microstructure signals to support short-horizon decision making under extreme volatility. However, existing trading systems struggle to jointly reason over noisy multi-source web evidence while maintaining robustness to rapid price shocks at sub-second timescales. The first challenge lies in synthesizing unstructured web content, social sentiment, and structured OHLCV signals into coherent and interpretable trading decisions without amplifying spurious correlations, while the second challenge concerns risk control, as slow deliberative reasoning pipelines are ill-suited for handling abrupt market shocks that require immediate defensive responses. To address these challenges, we propose \textsc{WebCryptoAgent}, an agentic trading framework that decomposes web-informed decision making into modality-specific agents and consolidates their outputs into a unified evidence document for confidence-calibrated reasoning. We further introduce a decoupled control architecture that separates strategic hourly reasoning from a real-time second-level risk model, enabling fast shock detection and protective intervention independent of the trading loop. Extensive experiments on real-world cryptocurrency markets demonstrate that \textsc{WebCryptoAgent} improves trading stability, reduces spurious activity, and enhances tail-risk handling compared to existing baselines.
Code and Demo: \url{https://github.com/AIGeeksGroup/WebCryptoAgent}.

\end{abstract}

\begin{figure}[t]
    \centering
    \includegraphics[width=\linewidth]{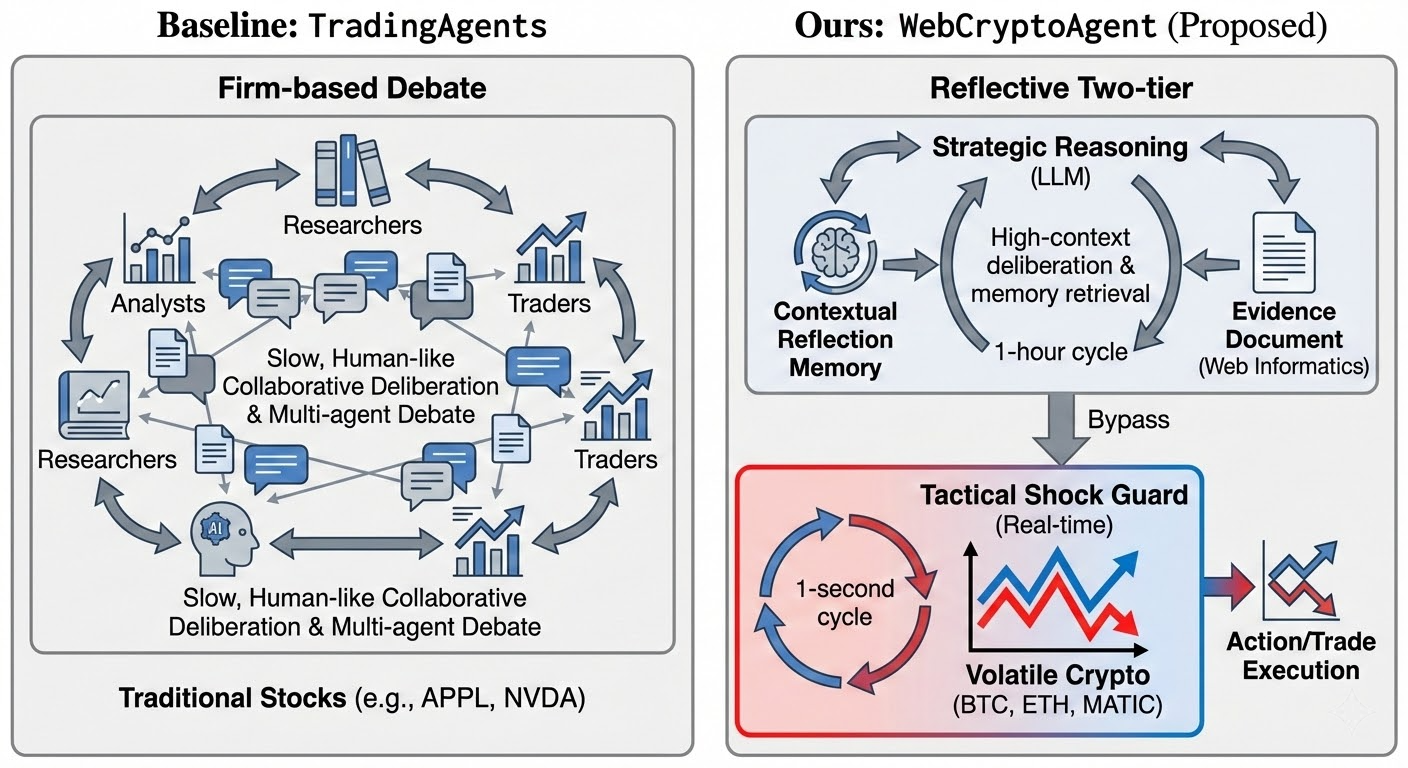}
    \caption{Structural comparison between the horizontal firm-based debate model (TradingAgents) and our proposed vertical reflective two-tier architecture (WebCryptoAgent).}
    \label{fig:arch_comparison}
\end{figure}

\section{Introduction}

In recent years, the rapid development of large language models (LLMs) has catalyzed a new paradigm of \emph{agentic trading systems}~\cite{shi2025presentagent,zhang2025marl,lin2025ccl,ge2025vasevqa,zhang2025vasevqa}, where autonomous agents leverage textual and numerical information to make financial decisions. With the global expansion of the cryptocurrency market, characterized by extreme volatility and round-the-clock trading, the demand for intelligent trading assistants has intensified. These agents are designed not only to process heterogeneous data sources—such as news, social media sentiment, and historical market data—but also to reason and act in dynamic environments. Early efforts in this direction include domain-adapted financial assistants such as PIXIU (FinMA)~\cite{xie2023pixiu}, FinGPT~\cite{yang2023fingpt}, and Instruct-FinGPT~\cite{zhang2023instructfingpt}, which fine-tune general-purpose LLMs on financial corpora to enhance domain sensitivity. Meanwhile, large-scale pretrained models such as BloombergGPT~\cite{wu2023bloomberggpt}, XuanYuan 2.0~\cite{zhang2023xuanyuan}, and Fin-T5~\cite{lu2023fint5} have demonstrated that hybrid domain–general corpora can achieve competitive reasoning capabilities while maintaining financial expertise. Collectively, these advances reveal the potential of language-based agents in financial contexts; however, most existing systems emphasize domain adaptation over agentic autonomy, leaving open challenges in continuous reasoning, contextual awareness, and decision self-correction.  

Beyond static financial modeling, recent work has explored LLM-based agents that directly interact with live trading environments. GPT-3.5/4 and open-source alternatives such as Qwen~\cite{bai2023qwen} and Baichuan~\cite{yang2023baichuan} have been tested on sentiment-driven trading tasks~\cite{lopez2023chatgpt}, showing promising profit margins even under naïve strategies. FinGPT-based pipelines~\cite{kirtac2024sentiment} and reasoning-augmented frameworks like WallStreetLLM~\cite{fatouros2024wall} extend this idea by incorporating news summarization and contextual interpretation. FinMem~\cite{yu2023finmem} and TradingGPT~\cite{li2023tradinggpt} introduce memory-enhanced and multi-agent debate mechanisms that reduce hallucination and improve backtesting performance, while hybrid RL-reflection designs such as SEP~\cite{koa2024sep} and PPO-augmented approaches~\cite{ding2023ppo} aim to optimize long-term trading returns. The latest evolution, TradingAgents~\cite{xiao2024tradingagents}, simulates an entire virtual trading firm where specialized LLM agents (analysts, researchers, traders, and risk managers) collaborate to achieve superior Sharpe ratios and drawdown control.

As illustrated in Figure~\ref{fig:arch_comparison}, while TradingAgents relies on a horizontal organizational structure with multiple specialized roles engaging in deliberative debate, our proposed \textsc{WebCryptoAgent} introduces a vertical, two-tier architecture specifically designed for the high-velocity requirements of cryptocurrency markets. This separation of strategic reasoning and tactical execution allows for complex decision-making without compromising the reaction speed necessary for crypto assets. Nevertheless, despite these innovations, two key challenges remain prevalent across agentic trading systems: (\emph{i}) limited self-correction capability, as current agents rarely utilize retrieved historical reasoning traces for reflective improvement; and (\emph{ii}) insufficient or underdeveloped risk management mechanisms, leading to unstable performance in volatile crypto markets.

Motivated by these observations, we aim to address the aforementioned limitations by introducing a novel agentic architecture that integrates \emph{contextual reflection} and \emph{structured risk management} into a unified pipeline. Our motivation stems from two core needs: first, enabling trading agents to autonomously reflect on past reasoning trajectories, refine decision policies, and adapt to evolving market conditions; and second, embedding robust risk assessment and control procedures into the decision loop to ensure both profitability and stability in high-risk environments such as cryptocurrency trading. By combining reflective reasoning with dynamic risk calibration, our approach aspires to move beyond single-step prediction toward sustained, self-corrective intelligence.

To realize these goals, we propose \textbf{WebCryptoAgent}, an end-to-end web-enabled crypto trading agent designed to perform autonomous trading, self-reflection, and adaptive risk management. Specifically, we design a \emph{contextual reflection module} that leverages retrieved decision histories and environmental cues to iteratively refine policy reasoning. In parallel, we introduce a \emph{hierarchical risk management framework} that evaluates portfolio exposure, volatility dynamics, and model uncertainty to adjust position sizes and safeguard returns. Furthermore, we conduct comprehensive experiments across multiple benchmark datasets and real-world simulation environments, demonstrating that WebCryptoAgent consistently outperforms existing baselines in profitability, stability, and drawdown control.

In summary, our main contributions can be outlined as follows:
\begin{itemize}
    \item \textbf{WebCryptoAgent Framework:} We introduce an agentic trading pipeline that integrates reasoning, self-reflection, and execution for cryptocurrency markets. The proposed contextual reflection module enables dynamic policy refinement based on historical feedback.
    \item \textbf{Hierarchical Risk Management:} We design a multi-level risk assessment mechanism (as shown in the ``Tactical Shock Guard'' of Figure~\ref{fig:arch_comparison}) that quantifies uncertainty, manages portfolio exposure, and prevents excessive drawdowns in high-volatility environments.
    \item \textbf{Comprehensive Evaluation:} Through extensive experiments on synthetic and real-world crypto datasets, we show that WebCryptoAgent achieves superior performance in cumulative return, Sharpe ratio, and risk-adjusted metrics compared to state-of-the-art agentic traders.
\end{itemize}

\section{Related Work}

\paragraph{Agentic Financial Assistants}
Domain-adapted language models for finance are generally obtained either through fine-tuning general-purpose LLM agents or pretraining from scratch on financial corpora. Fine-tuning enhances a model’s domain sensitivity while retaining its general reasoning ability. Examples include PIXIU (FinMA)~\cite{xie2023pixiu}, which fine-tunes LLaMA on 136K finance-related instructions; FinGPT~\cite{yang2023fingpt}, which applies LoRA to models such as LLaMA and ChatGLM with roughly 50K finance-specific samples; and Instruct-FinGPT~\cite{zhang2023instructfingpt}, which incorporates 10K sentiment-oriented instruction datasets. These specialized variants significantly outperform untuned models like BLOOM or OPT~\cite{zhang2022opt} on classification benchmarks, sometimes even surpassing BloombergGPT~\cite{wu2023bloomberggpt}, though they typically fall short of GPT-4 on open-ended reasoning tasks.  
Another line of work trains finance-specific LLM agents entirely from scratch. BloombergGPT~\cite{wu2023bloomberggpt}, XuanYuan 2.0~\cite{zhang2023xuanyuan}, and Fin-T5~\cite{lu2023fint5} exemplify this trend, using mixtures of general text and finance-domain corpora. BloombergGPT, in particular, demonstrates superior performance on market sentiment classification while remaining competitive on general NLP tasks. Collectively, these studies highlight the value of high-quality domain corpora in adapting LLM agents to financial contexts.

\paragraph{Agentic Traders}
LLM agents have also been positioned as autonomous trading agents capable of ingesting heterogeneous market signals and issuing trading actions. News-driven agents rely on textual market updates, financial reports, and sentiment analysis. Both closed-source models (e.g., GPT-3.5/4) and open-source LLMs (e.g., Qwen~\cite{bai2023qwen}, Baichuan~\cite{yang2023baichuan}) have been tested on stock-news sentiment prediction~\cite{lopez2023chatgpt}, with even simple sentiment-based strategies producing nontrivial returns. Further improvements arise from fine-tuned variants such as FinGPT or OPT-based financial sentiment models~\cite{kirtac2024sentiment}, as well as reasoning-augmented pipelines that summarize and interpret evolving news streams~\cite{fatouros2024wall}.  
Beyond direct sentiment mapping, reasoning-enhanced frameworks such as FinMem~\cite{yu2023finmem} integrate layered memory to contextualize decisions, while TradingGPT~\cite{li2023tradinggpt} employs multi-agent debates with distinct agent profiles. Such designs reduce hallucinations and yield superior backtest metrics. Reinforcement learning methods further refine trading performance by optimizing outputs against simulated returns; SEP~\cite{koa2024sep} exemplifies this reflection–RL hybrid, while PPO-based approaches~\cite{ding2023ppo} integrate LLM-generated embeddings into conventional RL pipelines. Recent work such as TradingAgents~\cite{xiao2024tradingagents} extends this direction by simulating a realistic trading firm environment with multiple specialized LLM agents (analysts, researchers, traders, and risk managers), achieving superior cumulative returns, Sharpe ratios, and drawdown control compared to traditional baselines.

\paragraph{Agentic Alpha Miners}
Instead of executing trades, LLM agents can also contribute by generating \emph{alpha factors}, i.e., novel predictive signals for trading. QuantAgent~\cite{wang2024quantagent} demonstrates a nested loop design in which a writer agent proposes scripts for factor generation, a judge agent provides feedback, and outer-loop evaluation against market data closes the feedback cycle. AlphaGPT~\cite{wang2023alphagpt} extends this to a human-in-the-loop paradigm where experts collaborate with agents to iteratively refine alpha strategies. These systems underscore the potential of LLM-driven alpha discovery, highlighting their ability to automate exploratory research and accelerate quantitative investment strategy design.

\section{Method}

\begin{figure*}[t]
  \centering
  \includegraphics[width=\linewidth]{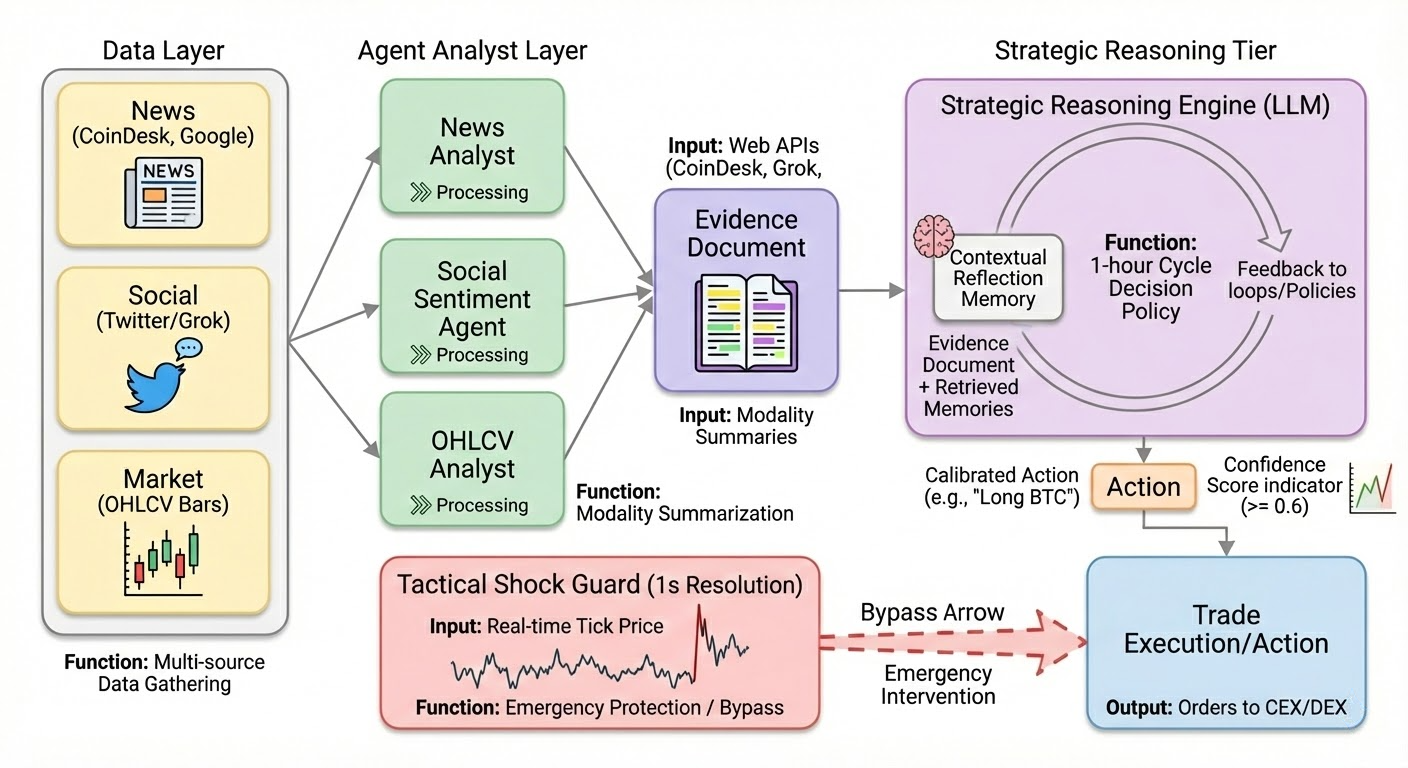}
  \caption{Overview of the WebCryptoAgent architecture. The framework employs a two-tier decision-making process: (1) a Strategic Tier where specialized agents aggregate multi-modal data (News, Social, Market) into an Evidence Document for LLM-based reasoning with contextual memory reflection; and (2) a Tactical Tier (Shock Guard) that monitors high-frequency tick data to trigger low-latency emergency bypasses. Final actions are dispatched to the Execution Layer for CEX/DEX deployment.}
  \label{fig:webcryptoagent-workflow}
\end{figure*}

\subsection{Overview}

Our approach integrates large language model (LLM) reasoning with systematic trading execution through three interdependent components:
(1) an agentic reasoning workflow for multi-modal market understanding,
(2) a contextual reflection mechanism inspired by Reflexion ~\citep{liu-etal-2025-contextual}, and
(3) a regime-aware risk management layer ensuring capital efficiency and adaptive exposure.

\subsection{Agent Workflow}

The proposed trading agent operates as a reasoning–execution pipeline that transforms heterogeneous market inputs into structured trading decisions. 
At each decision epoch $t$, the agent constructs a market snapshot 
$\mathcal{D}_t = \{ O_t, I_t, N_t, R_t \}$, 
where $O_t$ denotes multi-scale OHLCV data (15-minute and 1-hour bars), and
$I_t$ represents the indicator set
\[
\mathcal{I} \coloneqq
\textstyle\left\{
\begin{aligned}
&\mathrm{EMA}_{21}, \mathrm{EMA}_{50}, \mathrm{EMA}_{200}, \mathrm{RSI}_{14}, \\
&\mathrm{MACD}, \mathrm{ATR}_{14}, \mathrm{BB}, \mathrm{VWAP}, \mathrm{PDH}, \mathrm{PDL}
\end{aligned}
\right\}.
\]

encodes the current regime snapshot describing macro sentiment, volatility state, and liquidity depth.

Before decision generation, the agent retrieves contextually similar historical episodes from the experience memory $\mathcal{B}$ through a top-$K$ similarity search:
\[
\mathcal{E}_t = \mathrm{TopK}(\mathcal{B}, \mathcal{D}_t, K),
\]
where similarity is defined by a weighted combination of cosine distance in embedding space and exact regime matching.  
This retrieved context provides exemplars of how analogous market states evolved in the past.

The reasoning model $f_{\mathrm{LLM}}(\cdot)$, implemented using a large-language-model backbone (e.g., GPT-5 or Gemini-2.0-Flash-Thinking), processes both the current context and retrieved experiences to generate a structured decision tuple:
\[
\mathcal{A}_t = f_{\mathrm{LLM}}(\mathcal{D}_t, \mathcal{E}_t, R_t) 
= \{ b_t, c_t, m_t, \rho_t \},
\]
where $b_t \in \{\mathrm{LONG}, \mathrm{FLAT}\}$ is the directional bias, 
$c_t \in [0,1]$ is the confidence score, 
$m_t$ is the expected move in basis points, 
and $\rho_t$ is the generated rationale explaining the recognized pattern.

To avoid unstable oscillations in trade direction, we employ a regime-dependent hysteresis function:
\[
b_t =
\begin{cases}
\mathrm{LONG}, &
\begin{aligned}
c_t \, p_{\mathrm{long}} &\ge \theta_{\mathrm{adopt}}(R_t),\\
&\text{trigger fired},
\end{aligned}
\\[4pt]
\mathrm{FLAT}, &
c_t \, p_{\mathrm{long}} < \theta_{\mathrm{hold}}(R_t),
\\[4pt]
b_{t-1}, & \text{otherwise}.
\end{cases}
\]

Thresholds $\theta_{\mathrm{adopt}}$ and $\theta_{\mathrm{hold}}$ are adaptively calibrated by regime type, with $\theta_{\mathrm{adopt}} > \theta_{\mathrm{hold}}$ to enforce persistence.  
A bias refresh occurs every eight hours, ensuring adaptation to new regimes while maintaining temporal stability.

The overall strategic decision process is summarized in
Algorithm~\ref{alg:agent-workflow}.

\begin{algorithm}[t]
\caption{Strategic Agent Decision Workflow}
\label{alg:agent-workflow}
\KwIn{Market data streams at time $t$, replay buffer $\mathcal{B}$}
\KwOut{Trading action $a_t$}

Construct market snapshot $\mathcal{D}_t = \{O_t, I_t, N_t, R_t\}$\;

Retrieve contextual experiences $\mathcal{E}_t \leftarrow \mathrm{TopK}(\mathcal{B}, \mathcal{D}_t, K)$\;

Generate decision tuple
$\mathcal{A}_t = \{b_t, c_t, m_t, \rho_t\}
\leftarrow f_{\mathrm{LLM}}(\mathcal{D}_t, \mathcal{E}_t, R_t)$\;

Update directional bias via regime-dependent hysteresis (Eq.~(H))\;

\If{$c_t \ge \theta_{\mathrm{exec}}(R_t)$}{
    Execute trade with size determined by risk controller\;
}
\Else{
    Abstain from trading\;
}

\Return{$a_t$}
\end{algorithm}

\vspace{1em}

\subsection{Contextual Reflection}

Our self-improvement process is inspired by the \textit{Reflexion} framework~\citep{shinn2023reflexion} and extended through \textit{Contextual Experience Replay (CER)}~\citep{liu-etal-2025-contextual}. 
This component allows the agent to iteratively evaluate its own decisions, identify sources of error, and incorporate refined insights back into its reasoning context.

After each trade cycle, the agent observes realized outcomes at multiple horizons (4h, 8h, 24h, 7d) and forms a post-trade tuple:
\[
\tau_t = (\mathcal{D}_t, \mathcal{A}_t, r_{h,t}),
\]
where $r_{h,t}$ is the realized net return (in basis points) after transaction costs.  
A reflection query is then composed for the LLM, containing the trade rationale $\rho_t$, the corresponding outcomes, and the regime context at entry.  
The LLM outputs a structured reflection:
\[
\mathcal{F}_t \coloneqq
\textstyle\left\{
\begin{aligned}
&\text{outcome\_label},\; \text{attribution},\\
&\text{lesson},\; \text{pattern\_validity}
\end{aligned}
\right\}.
\]
where the outcome label $\in \{\mathrm{WIN}, \mathrm{LOSS}, \mathrm{BREAK\_EVEN}\}$ and the attribution field explains which input signals (technical, news, regime) most contributed to performance.

Each reflection is distilled into a compressed experience embedding:
\[
e_t \coloneqq \mathrm{Distill}(\tau_t)
= \textstyle\left\{
\begin{aligned}
&\text{context}_{\mathrm{embed}},\; R_t,\; \text{pattern},\\
&\text{cost},\; \{r_h\},\; \text{lesson}
\end{aligned}
\right\}.
\]
which is stored in the replay buffer $\mathcal{B}$ with exponential decay 
$w(e_t,t') = \exp(-\frac{t'-t}{\lambda})$, 
where $\lambda$ is the half-life parameter (e.g., 30 days).  
During future inference cycles, the agent retrieves top-$K$ semantically similar experiences from $\mathcal{B}$ and conditions the next reasoning step on these reflections, effectively reusing its prior knowledge as contextual exemplars.

This closed reflection–replay loop enables continual self-improvement without retraining.  
Over time, the agent develops regime-specific priors on success likelihoods and adaptively modifies its decision thresholds based on accumulated experience.  
Empirically, this feedback mechanism increases consistency, reduces regime-specific overconfidence, and leads to smoother cumulative performance trajectories.

The contextual reflection and experience replay mechanism is formalized
in Algorithm~\ref{alg:reflection}.

\begin{algorithm}[t]
\caption{Contextual Reflection and Experience Replay (CER)}
\label{alg:reflection}
\KwIn{Executed trade $\mathcal{A}_t$, realized returns $\{r_{h,t}\}$}
\KwOut{Updated replay buffer $\mathcal{B}$}

Form post-trade tuple $\tau_t = (\mathcal{D}_t, \mathcal{A}_t, r_{h,t})$\;

Query LLM for structured reflection
$\mathcal{F}_t \leftarrow \mathrm{Reflect}(\tau_t)$\;

Distill compressed experience embedding
$e_t \leftarrow \mathrm{Distill}(\tau_t, \mathcal{F}_t)$\;

Assign decay weight $w(e_t) = \exp(-\frac{t'-t}{\lambda})$\;

Insert $(e_t, w)$ into replay buffer $\mathcal{B}$\;

\Return{$\mathcal{B}$}
\end{algorithm}

\subsection{Risk Management}

The risk management subsystem converts qualitative reasoning outputs into executable, quantitatively constrained trades.
Position sizing is based on Average True Range (ATR)–derived volatility measures, where the stop-distance multiplier adapts to the current regime. In stable RISK-ON phases, positions are larger and stops tighter; during high-volatility or RISK-OFF periods, exposure is reduced and stops widened.
Position sizes are further modulated using a fractional Kelly criterion, linking LLM confidence to statistical edge estimation while capping leverage through a conservative scaling factor.
To ensure capital preservation, a hierarchy of risk controls is applied:
\begin{itemize}
    \item Circuit breakers halt trading after predefined loss or drawdown thresholds.
    \item Portfolio exposure limits restrict concentration by asset and by total equity share.
    \item Time-based stops close positions automatically when liquidity deteriorates or when maximum holding durations are reached.
\end{itemize}
Before order submission, an explicit cost gate compares the model’s expected edge against cumulative frictional costs (liquidity-provider fee, impact, gas, spread, and MEV). Trades are executed only if the expected return exceeds the estimated cost margin.

The overall end-to-end operation of WebCryptoAgent is summarized in
Algorithm~\ref{alg:overall}.

\begin{algorithm}[t]
\caption{Overall WebCryptoAgent Pipeline}
\label{alg:overall}
\KwIn{Streaming market data, web signals, replay buffer $\mathcal{B}$}
\KwOut{Executed trades and updated memory}

\While{market is open}{
    Collect multi-source inputs (News, Social, OHLCV)\;

    \tcp{Strategic Tier (hourly cadence)}
    \If{decision epoch reached}{
        Generate trading action $a_t$ via Strategic Agent (Algorithm~\ref{alg:agent-workflow})\;
    }

    \tcp{Tactical Tier (second-level monitoring)}
    Monitor high-frequency price stream for shock conditions\;
    \If{shock detected}{
        Override strategic action and trigger emergency protection\;
    }

    \tcp{Execution}
    Submit final action to execution layer (CEX/DEX)\;

    \tcp{Post-trade reflection}
    \If{trade cycle completed}{
        Update replay buffer $\mathcal{B}$ via Contextual Reflection (Algorithm~\ref{alg:reflection})\;
    }
}
\Return{Executed trades and updated replay buffer $\mathcal{B}$}
\end{algorithm}

\section{Experiment}
This section reports the empirical performance of four LLM-based trading agents on BTCUSDT, evaluated with and without memory. All results are produced under identical market data, execution rules, and decision schedules.

\subsection{Experimental Setting}
The experiment is conducted on BTCUSDT using 15-minute OHLCV data from 2025-01-05 to 2026-01-05, totaling 35,040 bars.  
Each model generates trading decisions at 122 fixed timestamps. Position sizing, transaction logic, and initial equity (\$10,000) are held constant across all runs.

Two configurations are evaluated:
\begin{itemize}
    \item \textbf{Memory-enabled:} the model receives past decision–outcome information.
    \item \textbf{No-memory:} the model acts solely on the current market snapshot.
\end{itemize}

\subsection{Cumulative Return}
\begin{figure}[t]
    \centering
    \includegraphics[width=\linewidth]{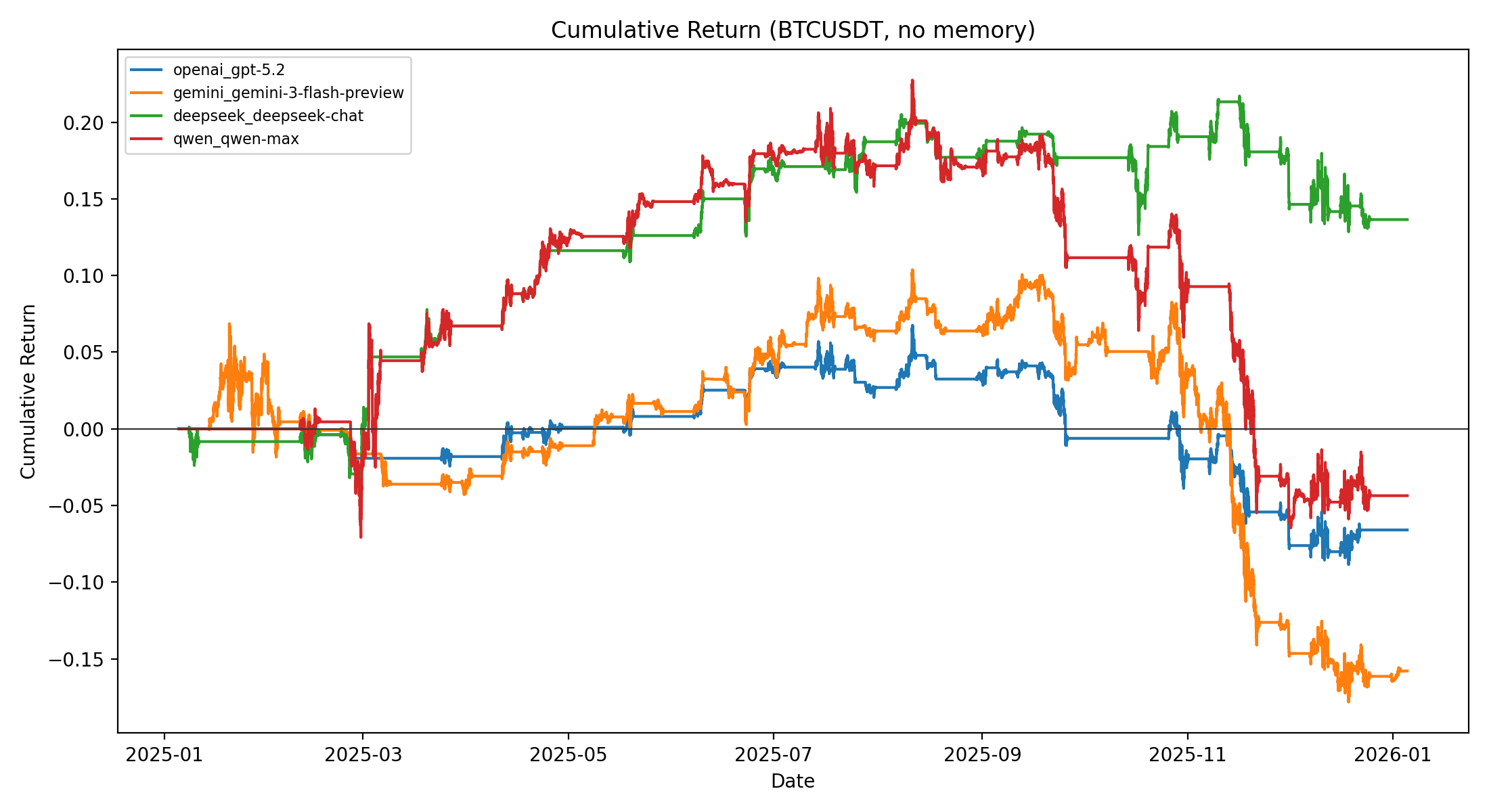}
    \includegraphics[width=\linewidth]{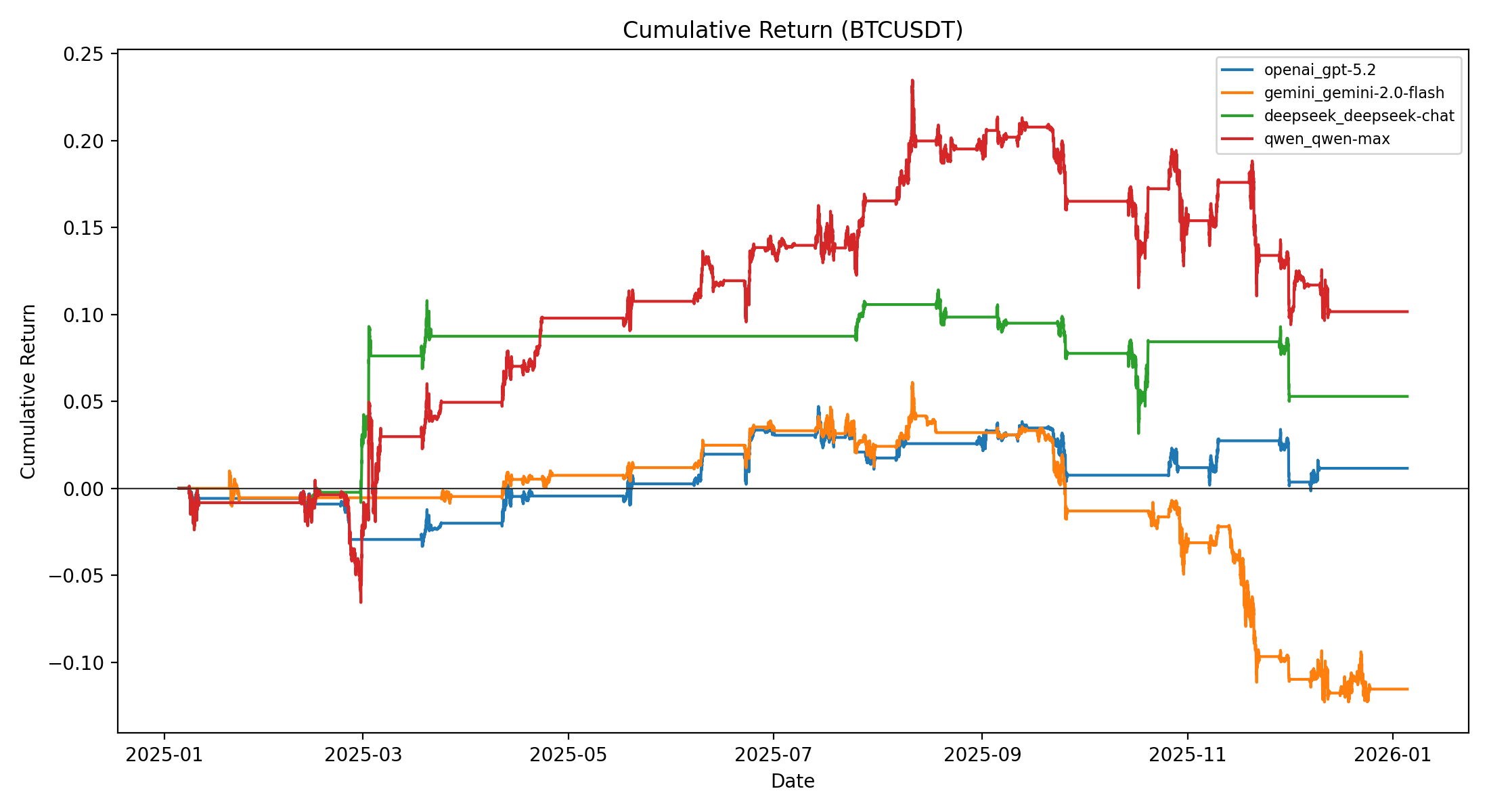}
    \caption{
    Cumulative return on BTCUSDT from 2025-01-05 to 2026-01-05.
    \textbf{Top:} no-memory configuration.
    \textbf{Bottom:} memory-enabled configuration.
    Each line corresponds to one LLM trading agent.
    }
    \label{fig:equity_compare}
\end{figure}

Figure~\ref{fig:equity_compare} shows cumulative return curves for all models under both configurations.

The figure shows visible differences in return trajectories, drawdowns, and final equity between models and between memory settings.

\subsection{BTCUSDT Results}
Table~\ref{tab:results} reports summary statistics for all runs, including total return, drawdown, Sharpe ratio, and final equity.
\begin{table*}[t]
\centering
\small
\begin{tabular}{l l r r r r r r}
\hline
Model & Memory & Trades & Win Rate & Total Ret. & Max DD & Sharpe & Equity End \\
\hline
GPT-5.2 & On  & 23 & 0.61 & 0.0115 & 0.0464 & 0.21 & 10115 \\
GPT-5.2 & Off & 27 & 0.56 & -0.0659 & 0.1461 & -0.67 & 9341 \\
Gemini Flash & On  & 26 & 0.42 & -0.1155 & 0.1732 & -1.27 & 8845 \\
Gemini Flash & Off & 50 & 0.46 & -0.1579 & 0.2553 & -0.89 & 8421 \\
DeepSeek Chat & On  & 10 & 0.50 & 0.0529 & 0.0742 & 0.76 & 10529 \\
DeepSeek Chat & Off & 29 & 0.66 & 0.1365 & 0.0728 & 1.19 & 11365 \\
Qwen-Max & On  & 36 & 0.64 & 0.1016 & 0.1139 & 0.80 & 11016 \\
Qwen-Max & Off & 42 & 0.62 & -0.0436 & 0.2378 & -0.17 & 9564 \\
\hline
\end{tabular}
\caption{Performance metrics for BTCUSDT trading experiments with and without memory.}
\label{tab:results}
\end{table*}

\begin{figure}[t]
    \centering
    \begin{subfigure}{\linewidth}
        \centering
        \includegraphics[width=\linewidth]{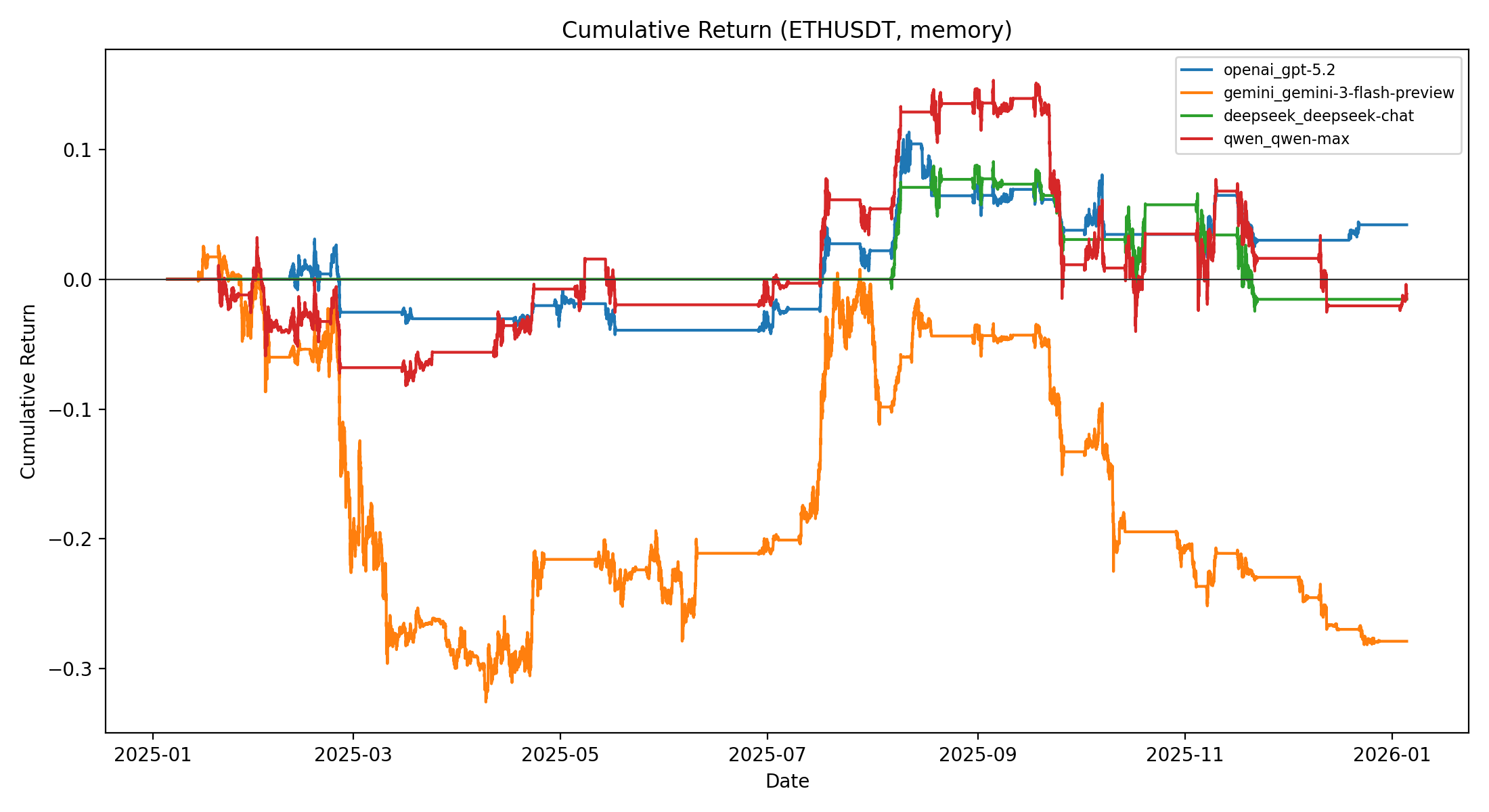}
        \caption{Cumulative return for ETHUSDT with memory enabled.}
        \label{fig:eth_equity_memory}
    \end{subfigure}

    \vspace{0.5em}

    \begin{subfigure}{\linewidth}
        \centering
        \includegraphics[width=\linewidth]{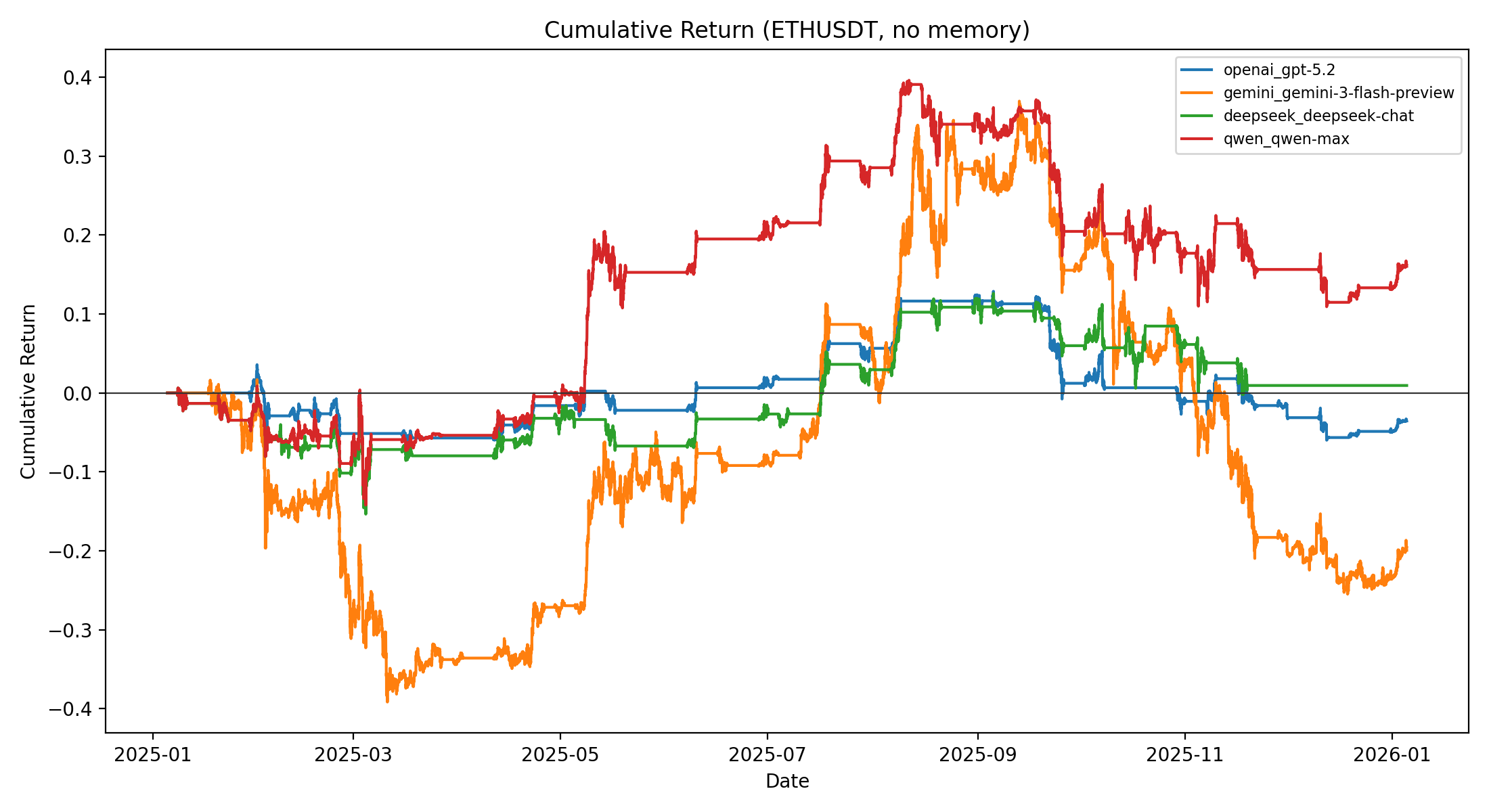}
        \caption{Cumulative return for ETHUSDT without memory.}
        \label{fig:eth_equity_nomemory}
    \end{subfigure}

    \caption{Equity curves for ETHUSDT trading with and without contextual memory.}
    \label{fig:eth_equity_compare}
\end{figure}

\subsection{ETHUSDT Results (Memory vs No-Memory)}

We repeat the same evaluation protocol on ETHUSDT over 2025-01-05 to 2026-01-05 using 15-minute bars (35,040 bars) and 122 decision points. Table~\ref{tab:eth_results} summarizes performance for each model under memory-enabled and no-memory configurations.

Overall, the results differ across model backbones and between memory settings. GPT-5.2 shifts from a negative return without memory to a positive return with memory. DeepSeek-Chat changes from a small positive return without memory to a small negative return with memory. Qwen-Max shows the opposite pattern, achieving its strongest performance in the no-memory configuration, while memory reduces its return.

\begin{table*}[t]
\centering
\small
\resizebox{\linewidth}{!}{%
\begin{tabular}{l l r r r r r r r r r r r}
\hline
Provider & Model & Memory & Trades & Win Rate & Total Ret. & CAGR & Max DD & Sharpe & Avg Ret/Trade & Median Ret/Trade & Equity End & Fallbacks \\
\hline
openai & gpt-5.2 & On  & 26 & 0.5769 & 0.0419 & 0.0420 & 0.0868 & 0.4334 & 0.00589 & 0.00272 & 10418.95 & 0 \\
openai & gpt-5.2 & Off & 32 & 0.5313 & -0.0355 & -0.0356 & 0.1671 & -0.2246 & 0.00089 & 0.00272 & 9645.15 & 0 \\
\hline
gemini & gemini-3-flash-preview & On  & 50 & 0.3400 & -0.2788 & -0.2795 & 0.3425 & -0.9348 & -0.03063 & -0.02946 & 7211.94 & 4 \\
gemini & gemini-3-flash-preview & Off & 50 & 0.4000 & -0.1992 & -0.1997 & 0.4557 & -0.2579 & -0.02705 & -0.03442 & 8007.58 & 3 \\
\hline
deepseek & deepseek-chat & On  & 10 & 0.4000 & -0.0155 & -0.0155 & 0.1057 & -0.0973 & -0.01216 & -0.00754 & 9845.37 & 0 \\
deepseek & deepseek-chat & Off & 30 & 0.4667 & 0.00949 & 0.00951 & 0.1608 & 0.1468 & 0.00258 & -0.00021 & 10094.87 & 0 \\
\hline
qwen & qwen-max & On  & 34 & 0.5588 & -0.0148 & -0.0149 & 0.1678 & 0.0143 & 0.00066 & 0.00564 & 9851.64 & 0 \\
qwen & qwen-max & Off & 47 & 0.6383 & 0.1604 & 0.1609 & 0.2055 & 0.7325 & 0.02284 & 0.01820 & 11604.35 & 0 \\
\hline
\end{tabular}
}
\caption{ETHUSDT performance metrics for memory-enabled vs no-memory trading runs. All models are evaluated over the same period (2025-01-05 to 2026-01-05), using 15-minute bars (35,040) and 122 decision points.}
\label{tab:eth_results}
\end{table*}

\subsection{POLUSDT Results (Memory vs No-Memory)}

We evaluate LLM-based trading agents on POLUSDT over the period
2025-01-05 to 2026-01-05 using 15-minute OHLCV data (35,040 bars) and
122 fixed decision points.
All models operate under identical execution rules and initial equity
(\$10,000).

Table~\ref{tab:pol_results_combined} reports performance metrics for
memory-enabled runs (top) and no-memory runs (bottom).

\begin{table*}[t]
\centering
\small
\resizebox{\linewidth}{!}{%
\begin{tabular}{l l l r r r r r r r r r}
\hline
Provider & Model & Memory & Trades & Win Rate & Total Ret. & CAGR & Max DD & Sharpe & Avg Ret/Trade & Median Ret/Trade & Equity End \\
\hline
openai & gpt-5.2 & On  & 13 & 0.3846 & -0.0641 & -0.0643 & 0.1437 & -0.6121 & -0.03511 & -0.02342 & 9358.58 \\
gemini & gemini-3-flash-preview & On  & 34 & 0.2941 & -0.2868 & -0.2875 & 0.3201 & -1.3348 & -0.05613 & -0.06332 & 7131.94 \\
deepseek & deepseek-chat & On  & 0  & 0.0000 & 0.0000 & 0.0000 & 0.0000 & 0.0000 & 0.00000 & 0.00000 & 10000.00 \\
qwen & qwen-max & On  & 30 & 0.5333 & -0.0060 & -0.0060 & 0.2525 & 0.0949 & -0.01545 & 0.00916 & 9940.49 \\
\hline
openai & gpt-5.2 & Off & 21 & 0.4762 & -0.1571 & -0.1575 & 0.2389 & -0.7855 & -0.01778 & -0.00765 & 8429.40 \\
gemini & gemini-3-flash-preview & Off & 52 & 0.2692 & -0.4810 & -0.4819 & 0.5951 & -0.8395 & -0.06527 & -0.06772 & 5190.21 \\
deepseek & deepseek-chat & Off & 10 & 0.6000 & -0.0016 & -0.0016 & 0.1790 & 0.0693 & 0.00829 & 0.00951 & 9983.74 \\
qwen & qwen-max & Off & 39 & 0.3846 & -0.2912 & -0.2918 & 0.4191 & -0.8093 & -0.01801 & -0.01538 & 7088.23 \\
\hline
\end{tabular}
}
\caption{POLUSDT performance metrics for LLM-based trading agents with memory enabled (top) and without memory (bottom). All runs use the same time period and decision points.}
\label{tab:pol_results_combined}
\end{table*}

\begin{figure}[t]
    \centering
    \begin{subfigure}{\linewidth}
        \centering
        \includegraphics[width=\linewidth]{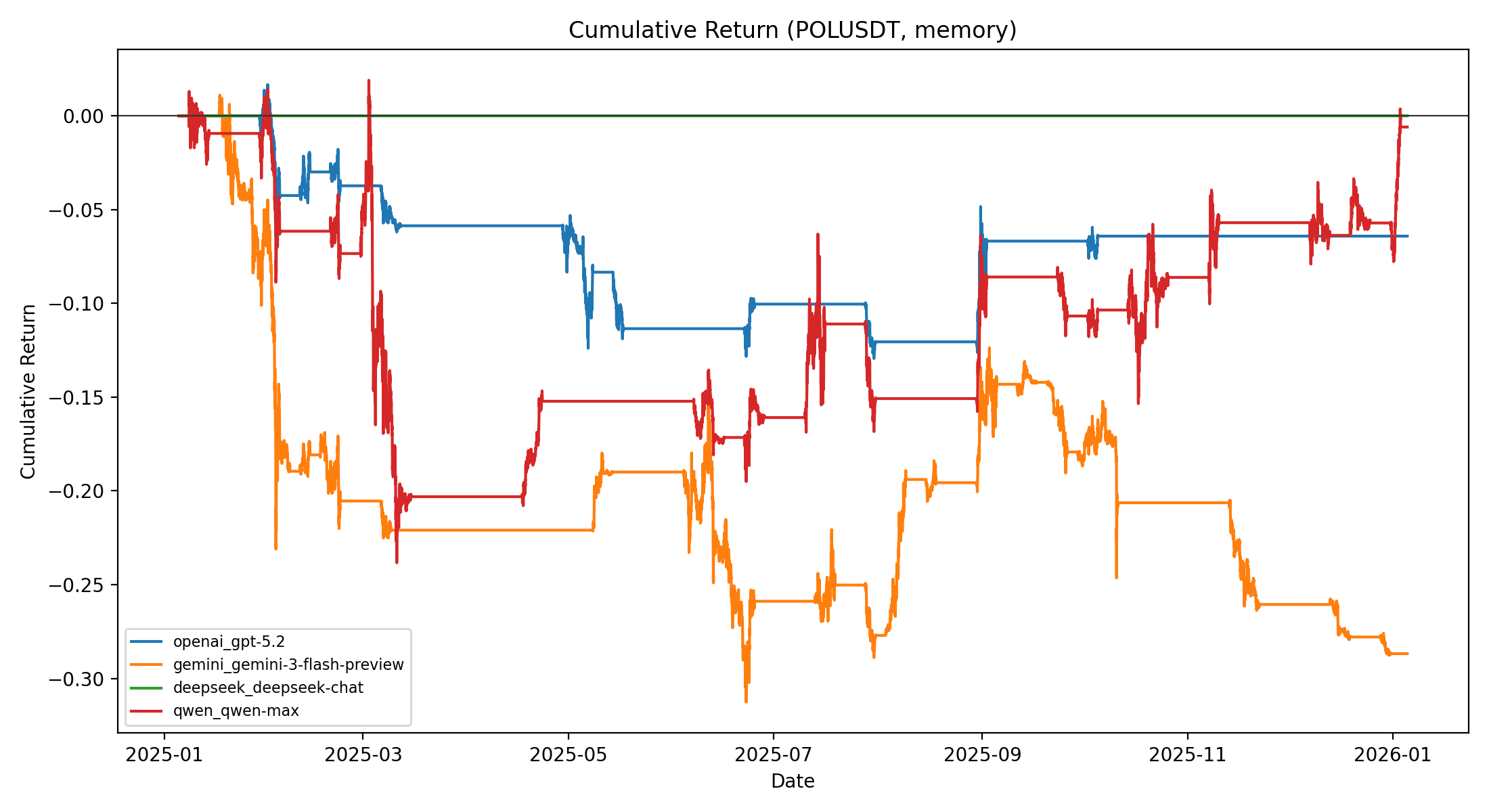}
        \caption{Cumulative return for POLUSDT with memory enabled.}
        \label{fig:pol_equity_memory}
    \end{subfigure}

    \vspace{0.5em}

    \begin{subfigure}{\linewidth}
        \centering
        \includegraphics[width=\linewidth]{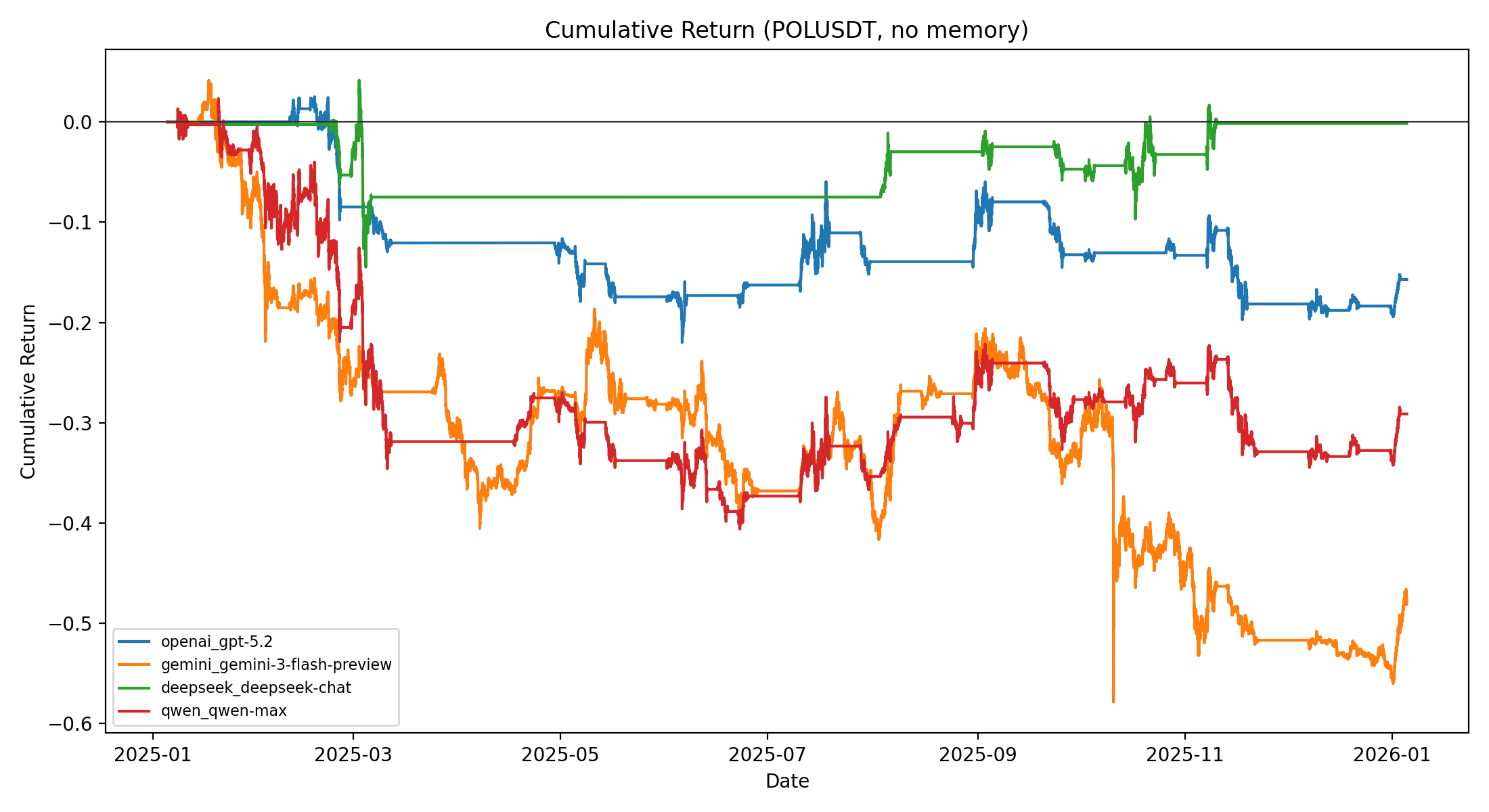}
        \caption{Cumulative return for POLUSDT without memory.}
        \label{fig:pol_equity_nomemory}
    \end{subfigure}

    \caption{Equity curves for POLUSDT trading with and without contextual memory.}
    \label{fig:pol_equity_compare}
\end{figure}

\subsection{ETHUSDT Equity Curves}

Figure~\ref{fig:eth_equity_memory} and Figure~\ref{fig:eth_equity_nomemory}
show cumulative returns for ETHUSDT with and without memory, evaluated over
the same time period and decision points as the BTCUSDT experiments.

\subsection{POLUSDT Equity Curves}

Figure~\ref{fig:pol_equity_memory} and Figure~\ref{fig:pol_equity_nomemory}
show cumulative returns for POLUSDT with and without memory under the same
evaluation protocol.

\subsection{Summary}
Across all models, both the cumulative return curves and summary metrics show that enabling memory leads to different performance outcomes compared to no-memory execution. The magnitude and direction of these differences vary across model backbones.

\section{Social Impact}

WebCryptoAgent illustrates how reflective, memory-augmented agentic systems can contribute to real-world financial infrastructures operating under extreme volatility. By decoupling strategic reasoning from low-latency risk control, the framework addresses the mismatch between deliberative decision making and the rapid dynamics of digital markets, enabling more stable and interpretable behavior. This design reduces excessive trading activity and mitigates abrupt losses, which is particularly relevant for retail participants and smaller institutions. Beyond individual performance, the contextual reflection and experience replay mechanism promotes adaptive yet conservative decision making without continuous retraining, allowing the agent to internalize regime-dependent priors and selectively abstain under uncertainty. Such behavior supports smoother trading dynamics and helps limit the amplification of noise-driven market fluctuations. At a broader level, WebCryptoAgent provides a practical blueprint for deploying large language models in high-stakes financial workflows where robustness and accountability are critical, and the two-tier reflective architecture may inform decision-support systems beyond cryptocurrency trading, including market monitoring and real-time economic analysis.

\section{Potential Risks}

The use of LLM-driven trading agents involves several practical considerations. Model behavior may vary under distribution shifts or rare market conditions, and reliance on external data sources can introduce noise or latency. In addition, reflection-based memory updates and automated execution require conservative configuration and ongoing monitoring. These considerations motivate cautious deployment and appropriate risk controls in real-world settings.

\section{Conclusion}

We presented WebCryptoAgent, a reflective agentic trading framework that integrates web-informed reasoning, contextual experience replay, and regime-aware risk control for short-horizon cryptocurrency trading. By decoupling strategic LLM-based reasoning from low-latency tactical protection, the proposed two-tier architecture enables robust decision making under extreme market volatility. Extensive experiments demonstrate that WebCryptoAgent improves trading stability, reduces spurious activity, and achieves stronger risk-adjusted performance compared to existing baselines. Beyond cryptocurrency markets, this work highlights the potential of reflective, memory-augmented agents for high-frequency decision-making tasks in dynamic and uncertain environments.

\section*{Limitation and Future Work}

While WebCryptoAgent demonstrates encouraging performance, several limitations remain. The framework currently relies on proprietary large language models for strategic reasoning, which may affect reproducibility across deployments. In addition, although the contextual reflection mechanism supports online adaptation without retraining, the replay buffer is updated using simple heuristics, and its long-term behavior warrants further study. Future work may explore alternative model choices, more principled reflection updates, and broader evaluation settings. We also expect that the two-tier reflective architecture could be applicable beyond cryptocurrency trading, though such extensions are left for future investigation.

\bibliography{custom}

@misc{xiao2024tradingagents,
      title={TradingAgents: Multi-Agents LLM Financial Trading Framework}, 
      author={Yijia Xiao and Edward Sun and Di Luo and Wei Wang},
      year={2025},
      eprint={2412.20138},
      archivePrefix={arXiv},
      primaryClass={q-fin.TR},
      url={https://arxiv.org/abs/2412.20138}, 
}

@misc{xie2023pixiu,
  title={PIXIU: A Large Language Model, Instruction Data and Evaluation Benchmark for Finance},
  author={Xie, Qianqian and Han, Weiguang and Zhang, Xiao and Lai, Yanzhao and Peng, Min and Lopez-Lira, Alejandro and Huang, Jimin},
  year={2023},
  eprint={2306.05443},
  archivePrefix={arXiv},
  primaryClass={cs.CL}
}

@misc{yang2023fingpt,
  title={FinGPT: Open-Source Financial Large Language Models},
  author={Yang, Hongyang and Liu, Xiao-Yang and Wang, Christina Dan},
  year={2023},
  eprint={2306.06031},
  archivePrefix={arXiv},
  primaryClass={q-fin.ST}
}

@misc{zhang2023instructfingpt,
  title={Instruct-FinGPT: Financial Sentiment Analysis by Instruction Tuning of General-Purpose Large Language Models},
  author={Zhang, Boyu and Yang, Hongyang and Liu, Xiao-Yang},
  year={2023},
  eprint={2306.12659},
  archivePrefix={arXiv},
  primaryClass={cs.CL}
}

@misc{zhang2022opt,
  title={OPT: Open Pre-trained Transformer Language Models},
  author={Zhang, Susan and Roller, Stephen and Goyal, Naman and Artetxe, Mikel and Chen, Moya and Chen, Shuohui and Dewan, Christopher and Diab, Mona and Li, Xian and Lin, Xi Victoria and Mihaylov, Todor and Ott, Myle and Shleifer, Sam and Shuster, Kurt and Simig, Daniel and Koura, Punit Singh and Sridhar, Anjali and Wang, Tianlu and Zettlemoyer, Luke},
  year={2022},
  eprint={2205.01068},
  archivePrefix={arXiv},
  primaryClass={cs.CL}
}

@misc{wu2023bloomberggpt,
  title={BloombergGPT: A Large Language Model for Finance},
  author={Wu, Shijie and Irsoy, Ozan and Lu, Steven and Dabravolski, Vadim and Dredze, Mark and Gehrmann, Sebastian and Kambadur, Prabhanjan and Rosenberg, David and Mann, Gideon},
  year={2023},
  eprint={2303.17564},
  archivePrefix={arXiv},
  primaryClass={cs.LG}
}

@misc{zhang2023xuanyuan,
  title={XuanYuan 2.0: A Large Chinese Financial Chat Model with Hundreds of Billions Parameters},
  author={Zhang, Xuanyu and Yang, Qing and Xu, Dongliang},
  year={2023},
  eprint={2305.12002},
  archivePrefix={arXiv},
  primaryClass={cs.CL}
}

@misc{lu2023fint5,
  title={BBT-Fin: Comprehensive Construction of Chinese Financial Domain Pre-trained Language Model, Corpus and Benchmark},
  author={Lu, Dakuan and Wu, Hengkui and Liang, Jiaqing and Xu, Yipei and He, Qianyu and Geng, Yipeng and Han, Mengkun and Xin, Yingsi and Xiao, Yanghua},
  year={2023},
  eprint={2302.09432},
  archivePrefix={arXiv},
  primaryClass={cs.CL}
}

@misc{bai2023qwen,
  title={Qwen Technical Report},
  author={Bai, Jinze and Bai, Shuai and Chu, Yunfei and Cui, Zeyu and Dang, Kai and Deng, Xiaodong and Fan, Yang and Ge, Wenbin and Han, Yu and Huang, Fei and Hui, Binyuan and Ji, Luo and Li, Mei and Lin, Junyang and Lin, Runji and Liu, Dayiheng and Liu, Gao and Lu, Chengqiang and Lu, Keming and Ma, Jianxin and Men, Rui and Ren, Xingzhang and Ren, Xuancheng and Tan, Chuanqi and Tan, Sinan and Tu, Jianhong and Wang, Peng and Wang, Shijie and Wang, Wei and Wu, Shengguang and Xu, Benfeng and Xu, Jin and Yang, An and Yang, Hao and Yang, Jian and Yang, Shusheng and Yao, Yang and Yu, Bowen and Yuan, Hongyi and Yuan, Zheng and Zhang, Jianwei and Zhang, Xingxuan and Zhang, Yichang and Zhang, Zhenru and Zhou, Chang and Zhou, Jingren and Zhou, Xiaohuan and Zhu, Tianhang},
  year={2023},
  eprint={2309.16609},
  archivePrefix={arXiv},
  primaryClass={cs.CL}
}

@misc{yang2023baichuan,
  title={Baichuan 2: Open Large-scale Language Models},
  author={Yang, Aiyuan and Xiao, Bin and Wang, Bingning and Zhang, Borong and Bian, Ce and Yin, Chao and Lv, Chenxu and Pan, Da and Wang, Dian and Yan, Dong and Yang, Fan and Deng, Fei and Wang, Feng and Liu, Feng and Ai, Guangwei and Dong, Guosheng and Zhao, Haizhou and Xu, Hang and Sun, Haoze and Zhang, Hongda and Liu, Hui and Ji, Jiaming and Xie, Jian and Dai, JunTao and Fang, Kun and Su, Lei and Song, Liang and Liu, Lifeng and Ru, Liyun and Ma, Luyao and Wang, Mang and Liu, Mickel and Lin, MingAn and Nie, Nuolan and Guo, Peidong and Sun, Ruiyang and Zhang, Tao and Li, Tianpeng and Li, Tianyu and Cheng, Wei and Chen, Weipeng and Zeng, Xiangrong and Wang, Xiaochuan and Chen, Xiaoxi and Men, Xin and Yu, Xin and Pan, Xuehai and Shen, Yanjun and Wang, Yiding and Li, Yiyu and Jiang, Youxin and Gao, Yuchen and Zhang, Yupeng and Zhou, Zenan and Wu, Zhiying},
  year={2023},
  eprint={2309.10305},
  archivePrefix={arXiv},
  primaryClass={cs.CL}
}

@misc{lopez2023chatgpt,
  title={Can ChatGPT Forecast Stock Price Movements? Return Predictability and Large Language Models},
  author={Lopez-Lira, Alejandro and Tang, Yuehua},
  year={2023},
  eprint={2304.07619},
  archivePrefix={arXiv},
  primaryClass={q-fin.ST}
}

@article{kirtac2024sentiment,
  title={Sentiment trading with large language models},
  author={Kirtac, Kemal and Germano, Guido},
  journal={Finance Research Letters},
  volume={62},
  pages={105227},
  year={2024},
  doi={10.1016/j.frl.2024.105227},
  publisher={Elsevier}
}

@article{fatouros2024wall,
  title={Can Large Language Models Beat Wall Street? Unveiling the Potential of AI in Stock Selection},
  author={Fatouros, Georgios and Metaxas, Konstantinos and Soldatos, John and Kyriazis, Dimosthenis},
  journal={Neural Computing and Applications},
  year={2024},
  doi={10.1007/s00521-024-10613-4},
  publisher={Springer}
}

@misc{yu2023finmem,
  title={FinMem: A Performance-Enhanced LLM Trading Agent with Layered Memory and Character Design},
  author={Yu, Yangyang and Li, Haohang and Chen, Zhi and Jiang, Yuechen and Li, Yang and Zhang, Denghui and Liu, Rong and Suchow, Jordan W. and Khashanah, Khaldoun},
  year={2023},
  eprint={2311.13743},
  archivePrefix={arXiv},
  primaryClass={q-fin.CP}
}

@misc{li2023tradinggpt,
  title={TradingGPT: Multi-Agent System with Layered Memory and Distinct Characters for Enhanced Financial Trading Performance},
  author={Li, Yang and Yu, Yangyang and Li, Haohang and Chen, Zhi and Khashanah, Khaldoun},
  year={2023},
  eprint={2309.03736},
  archivePrefix={arXiv},
  primaryClass={q-fin.PM}
}

@inproceedings{koa2024sep,
  title={Learning to Generate Explainable Stock Predictions using Self-Reflective Large Language Models},
  author={Koa, Kelvin J. L. and Ma, Yunshan and Ng, Ritchie and Chua, Tat-Seng},
  booktitle={Proceedings of the ACM Web Conference 2024 (WWW '24)},
  year={2024},
  doi={10.1145/3589334.3645611},
  publisher={ACM}
}

@misc{ding2023ppo,
  title={Integrating Stock Features and Global Information via Large Language Models for Enhanced Stock Return Prediction},
  author={Ding, Yujie and Jia, Shuai and Ma, Tianyi and Mao, Bingcheng and Zhou, Xiuze and Li, Liuliu and Han, Dongming},
  year={2023},
  eprint={2310.05627},
  archivePrefix={arXiv},
  primaryClass={cs.CL}
}

@misc{wang2024quantagent,
  title={QuantAgent: Seeking Holy Grail in Trading by Self-Improving Large Language Model},
  author={Wang, Saizhuo and Yuan, Hang and Ni, Lionel M. and Guo, Jian},
  year={2024},
  eprint={2402.03755},
  archivePrefix={arXiv},
  primaryClass={cs.AI}
}

@inproceedings{shi2025presentagent,
  title={Presentagent: Multimodal agent for presentation video generation},
  author={Shi, Jingwei and Zhang, Zeyu and Wu, Biao and Liang, Yanjie and Fang, Meng and Chen, Ling and Zhao, Yang},
  booktitle={Proceedings of the 2025 Conference on Empirical Methods in Natural Language Processing: System Demonstrations},
  pages={760--773},
  year={2025}
}

@inproceedings{zhang2025marl,
  title={MARL-MambaContour: Unleashing Multi-Agent Deep Reinforcement Learning for Active Contour Optimization in Medical Image Segmentation},
  author={Zhang, Ruicheng and Sun, Yu and Zhang, Zeyu and Li, Jinai and Liu, Xiaofan and Au, Hoi Fan and Guo, Haowei and Yan, Puxin},
  booktitle={Proceedings of the 33rd ACM International Conference on Multimedia},
  pages={7815--7824},
  year={2025}
}

@inproceedings{lin2025ccl,
  title={CCL: collaborative curriculum learning for sparse-reward multi-agent reinforcement learning via co-evolutionary task evolution},
  author={Lin, Yufei and Ye, Chengwei and Zhang, Huanzhen and Wang, Kangsheng and Xu, Linuo and Liu, Shuyan and Zhang, Zeyu},
  booktitle={International Conference on Intelligent Computing},
  pages={51--62},
  year={2025},
  organization={Springer}
}

@article{ge2025vasevqa,
  title={Vasevqa: Multimodal agent and benchmark for ancient greek pottery},
  author={Ge, Jinchao and Cheng, Tengfei and Wu, Biao and Zhang, Zeyu and Huang, Shiya and Bishop, Judith and Shepherd, Gillian and Fang, Meng and Chen, Ling and Zhao, Yang},
  journal={arXiv preprint arXiv:2509.17191},
  year={2025}
}

@article{zhang2025vasevqa,
  title={VaseVQA-3D: Benchmarking 3D VLMs on Ancient Greek Pottery},
  author={Zhang, Nonghai and Zhang, Zeyu and Wang, Jiazi and Zhao, Yang and Tang, Hao},
  journal={arXiv preprint arXiv:2510.04479},
  year={2025}
}

@misc{wang2023alphagpt,
  title={Alpha-GPT: Human-AI Interactive Alpha Mining for Quantitative Investment},
  author={Wang, Saizhuo and Yuan, Hang and Zhou, Leon and Ni, Lionel M. and Shum, Heung-Yeung and Guo, Jian},
  year={2023},
  eprint={2308.00016},
  archivePrefix={arXiv},
  primaryClass={q-fin.CP}
}

@inproceedings{liu-etal-2025-contextual,
    title = "Contextual Experience Replay for Self-Improvement of Language Agents",
    author = "Liu, Yitao  and
      Si, Chenglei  and
      Narasimhan, Karthik R  and
      Yao, Shunyu",
    editor = "Che, Wanxiang  and
      Nabende, Joyce  and
      Shutova, Ekaterina  and
      Pilehvar, Mohammad Taher",
    booktitle = "Proceedings of the 63rd Annual Meeting of the Association for Computational Linguistics (Volume 1: Long Papers)",
    month = jul,
    year = "2025",
    address = "Vienna, Austria",
    publisher = "Association for Computational Linguistics",
    url = "https://aclanthology.org/2025.acl-long.694/",
    doi = "10.18653/v1/2025.acl-long.694",
    pages = "14179--14198",
    ISBN = "979-8-89176-251-0"
}

@article{shinn2023reflexion,
  title={Reflexion: Language Agents with Verbal Reinforcement Learning},
  author={Shinn, Noah and Cassano, Federico and Berman, Edward and Gopinath, Ashwin and Narasimhan, Karthik and Yao, Shunyu},
  journal={arXiv preprint arXiv:2303.11366},
  year={2023},
  url={https://arxiv.org/abs/2303.11366},
  doi={10.48550/arXiv.2303.11366}
}

\label{sec:appendix}
\appendix

\end{document}